\pdfoutput=1

\documentclass[11pt]{article}

\usepackage[]{acl}

\usepackage{adjustbox}
\usepackage{times}
\usepackage{latexsym}
\usepackage{amsmath}
\usepackage{amssymb}
\usepackage{booktabs}
\usepackage{multirow}
\usepackage{todonotes}
\usepackage{xcolor}
\usepackage{xspace}

\usepackage[T1]{fontenc}

\usepackage[utf8]{inputenc}

\usepackage{microtype}

\newcommand{\mucsum}{\textsc{MUCSUM}\xspace}
\newcommand{\muc}{\textsc{MUC-4}\xspace}
\newcommand{\gray}[1]{\textcolor{gray}{#1}}

\title{Event-Keyed Summarization}

\author{William Gantt \quad Alexander Martin \quad Pavlo Kuchmiichuk \quad Aaron Steven White \\
  University of Rochester \\
  \texttt{\small{\{wgantt@cs|amart50@u|pkuchmii@ur|aaron.white@\}.rochester.edu}}}

\begin{document}
\maketitle
\begin{abstract}
We introduce \emph{event-keyed summarization} (EKS), a novel task that marries traditional summarization and document-level event extraction, with the goal of generating a contextualized summary for a \emph{specific} event, given a document and an extracted event structure. We introduce a dataset for this task, \mucsum, consisting of summaries of all  events in the classic \muc dataset, along with a set of baselines that comprises both  pretrained LM standards in the summarization literature, as well as larger frontier models. We show that ablations that reduce EKS to traditional summarization or structure-to-text yield inferior summaries of target events and that \mucsum is a robust benchmark for this task. Lastly, we conduct a human evaluation of both reference and model summaries, and provide some detailed analysis of the results.\footnote{\url{https://github.com/wgantt/eks}.}
\end{abstract}

\section{Introduction}
\label{sec:introduction}
Classically, event extraction (EE) aims to produce structured event representations from unstructured text. As early as the Message Understanding Conferences of the 1990s \citep{grishman-sundheim-1996-message}, the motivation for EE was fundamentally human-centric: a desire for adaptive systems that could ``respond to a user's information need'' \citep{grishman-2019-twenty}. Yet, the  majority of EE research focuses intensively on improving metrics on major benchmarks without considering how those improvements may be leveraged to better serve users.

Arguably, the most human-centric way to convey information about complex events is with summaries. Prominent summarization datasets such as CNN/Daily Mail \citep{nallapati-etal-2016-abstractive-v2}, XSUM \citep{narayan-etal-2018-dont}, and GigaWord \citep{rush-etal-2015-neural} rely on some conception of what is most salient to the \emph{average} reader. While this is appropriate for many use cases, it is much less so when a user has a \emph{specific} information need, as in EE.

\begin{figure}
    \centering
    \includegraphics[width=\linewidth]{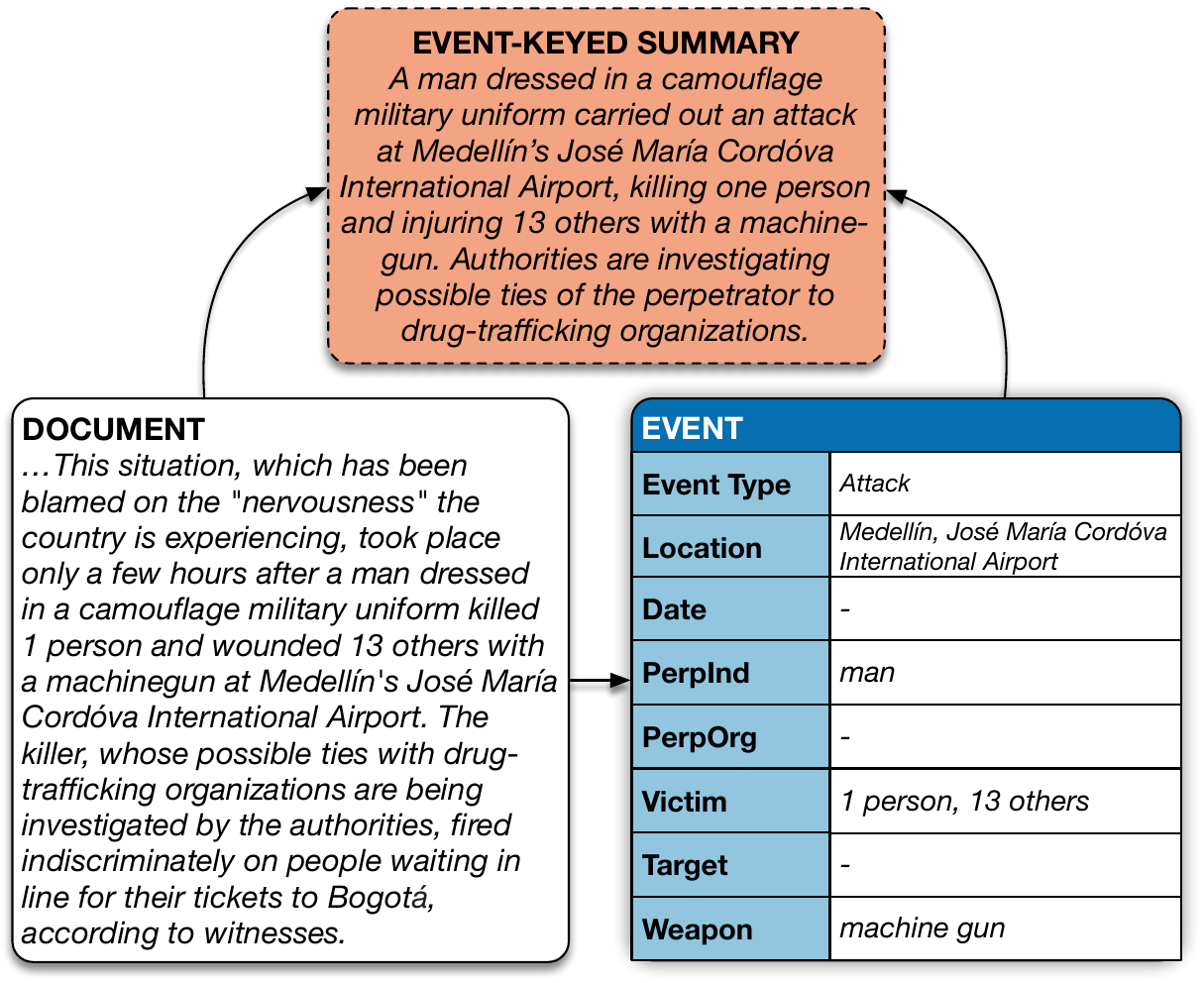}
    \caption{An illustration of the event-keyed summarization (EKS) task on a document and event template from the \mucsum training split. Given a document and event template, a system must generate a contextualized summary of that \emph{specific} event.\vspace{-8mm}}
    \label{fig:eks}
\end{figure}

Research on \emph{controllable} summarization tries to mediate this one-size-fits-all regime by giving users more command over particular summary attributes, such as length, style, or specificity \citep[][\emph{i.a.}]{fan-etal-2018-controllable, liu-etal-2018-controlling, he-etal-2022-ctrlsum, zhang-etal-2023-macsum}. Most relevant, \citet{s-hussain-etal-2022-event} use sets of extracted event keywords to guide generation of summaries covering multiple events. Our work is closely related but focuses on \emph{individual}, \emph{user-specified} events: we marry EE's focus on event-centric information needs with the user-friendliness of summarization to produce \emph{event-keyed summaries} (EKS): short, targeted summaries of a \emph{particular} event based on a document and an extracted event structure (\autoref{fig:eks}). We present \mucsum, a benchmark for EKS based on the classic MUC-4 dataset \citep{sundheim-1992-overview}, together with fine-tuned and zero-shot baselines, as well as ablations to show that our benchmark is \emph{not} readily reducible to either traditional summarization or structure-to-text formulations. Finally, we conclude with some analysis of a human evaluation of reference and model summaries.

\section{Task definition}
\label{sec:task-definition}
We define an \textit{event ontology} as a tuple $\langle\mathcal{E}, \mathcal{R}, \mathcal{S}\rangle$, consisting of a set of event types $\mathcal{E}$, a set of role types $\mathcal{R}$, and an assignment $\mathcal{S}: \mathcal{E} \rightarrow 2^\mathcal{R}$ of event types to sets of role types. We define an \textit{event} as a pair $\langle E, R \rangle$, consisting of an event type paired with a (possibly empty) set of \textit{event triggers} $E \in \mathcal{E} \times 2^{\Sigma^*}$ and a mapping from the roles associated with that event type to a (possibly empty) set of \textit{role fillers} $R: \mathcal{S}(E) \rightarrow 2^{\Sigma^*}$. 

We define \textit{event-keyed summarization} (EKS) as the task of mapping an input document $D \in \Sigma^*$ and a query event $\langle E, R \rangle$ to a summary $S \in \Sigma^*$ (with $\lvert S \rvert \ll \lvert D \rvert$) that conveys all and only the \emph{relevant} information in $D$ about $\langle E, R \rangle$---with relevance determined by the role set $\mathcal{S}(E)$.

\section{Data}
\label{sec:data}
\paragraph{MUC-4} We focus on the classic MUC-4 template extraction dataset as a case study \citep{muc-1992-message, sundheim-1992-overview}. In template extraction (contrasting with general event extraction), the set of event triggers is always empty, and so, in this case study, all information to be summarized comes from the event type and mapping from roles to role fillers.    

MUC-4 annotates 1,700 documents concerning political conflict in Latin American countries, with terrorism-focused event types $\mathcal{E} = $ \{\texttt{arson}, \texttt{attack}, \texttt{bombing}, \texttt{kidnapping}, \texttt{robbery}, \texttt{forced work stoppage}\}. All event types are associated with the same set of 24 roles---i.e. $\mathcal{S}[\mathcal{E}] = \mathcal{R}$, and each document may be associated with zero or more events of each type.

Since MUC-4, it has become standard to focus on a five-role subset, consisting of the individual perpetrator(s) (\texttt{PerpInd}), the organization(s) they are affiliated with (\texttt{PerpOrg}), the weapons they use (\texttt{Weapon}), victims of the incident (\texttt{Victim}), and damaged physical infrastructure (\texttt{Target}) \citep[][\emph{i.a.}]{chambers-jurafsky-2011-template, du-etal-2021-grit, chen-etal-2023-iterative, gantt-etal-2023-event, gantt2024multimuc}. We follow this practice here, but with two additions. First, we include the \texttt{StageOfExecution} role, which conveys whether the event actually occurred, was (unsuccessfully) attempted, or was merely threatened. Second, we include the \texttt{Location} and \texttt{Date} roles in cases where this information can actually be extracted from the text.\footnote{In MUC-4, arguments for \texttt{Location} and \texttt{Date} act more like document metadata, as in most cases their values are not actually extractable from the text itself. We (re-)annotate them only when they can be extracted.\label{fn:location_and_date}} These properties (time, location, and reality status) contain essential details about an event, and are necessary for a complete summary when provided.

\begin{table}[]
    \centering
    \begin{tabular}{l|ccc}
    \toprule
         & Train & Dev & Test \\
    \midrule
        Documents & 1,300 & \phantom{0}200 & \phantom{0}200 \\
        Events (summaries) & 1,114 & \phantom{0}191 & - \\
        Avg. words/doc & 328.5 & 354.1 & - \\
        Avg. sents/doc & \phantom{0}12.7 & \phantom{0}14.0 & - \\
        Avg. words/summary & \phantom{0}44.1 & \phantom{0}51.1 & - \\
        Avg. sents/summary & \phantom{00}1.7 & \phantom{00}1.8 & - \\
    \bottomrule
    \end{tabular}
    \caption{\mucsum dataset statistics. Detailed test set statistics are deliberately omitted.\vspace{-6mm}}
    \label{tab:dataset_statistics}
\end{table}

\paragraph{MUCSUM} Given the gold template annotations in MUC-4, we (three of the authors) wrote one \emph{abstractive} summary per document-template pair $\langle D, \langle E, R \rangle \rangle$ that aims to convey all relevant information about $\langle E, R \rangle$ provided in $D$, given the roles listed above. To reduce the burden of writing summaries for so many events, we adopted a generate-then-edit approach: we first prompted ChatGPT\footnote{\url{https://openai.com/blog/chatgpt.}} to produce a succinct ($\leq 3$-sentence) candidate summary conditioned on $\langle D, \langle E, R \rangle \rangle$, then manually edited the result to ensure that it contained all $\mathcal{R}$-relevant information represented in $D$ about $\langle E, R \rangle$. Additional information was included in the summaries if it provided important context or was otherwise necessary to ground the situation being described. In the course of writing summaries, we also re-annotated the \texttt{Date} and \texttt{Location} roles (see \autoref{fn:location_and_date}). Each $\langle D, \langle E, R \rangle \rangle$ pair was singly annotated, though we redundantly annotated a random subset of 30 test set examples and include agreement measures for these in \autoref{app:annotation}. We call the resulting dataset \textbf{MUCSUM} and provide summary statistics in \autoref{tab:dataset_statistics}.\footnote{Data and annotation instructions are in the GitHub repository. Details on the ChatGPT summarization prompt and hyperparameters are included in \autoref{app:models}.}

\begin{table*}[]
    \small{}
    \centering
    \begin{tabular}{ll|ccccccccc}
    \toprule
         Model & Setting & $\textbf{R}_1$ & $\textbf{R}_2$ & $\textbf{R}_L$ & \textbf{BS} & \textbf{CR} & $S_r \rightarrow S_p$ & $S_p \rightarrow S_r$ & $S_r \leftrightarrow S_p$ & $D \rightarrow S_p$\\
    \midrule
       3-Sent Baseline & - & 46.0 & 28.7 & 33.6 & 89.3 & 37.9 & \phantom{0}1.7 & \phantom{0}5.9 & \phantom{0}3.8 & \phantom{0}1.3 \\
       \midrule
       ChatGPT & temp+doc & 47.0 & 30.4 & 35.6 & 88.6 & 60.2 & 30.6 & \textbf{43.0} & 36.8 & 40.8 \\
       GPT-4 & temp+doc & 48.7 & 30.0 & 35.9 & 88.9 & 67.6 & \textbf{45.0} & 40.3 & \textbf{42.7} & 38.8 \\
    \midrule
       $\textsc{BART}$ & temp+doc & 66.7 & 47.9 & 52.7 & 93.4 & 71.8 & 39.9 & 30.6 & 35.2 & 37.4 \\ 
       & \gray{temp only} & \gray{51.9} & \gray{30.5} & \gray{37.9} & \gray{91.2} & \gray{74.6} & \gray{15.4} & \gray{\phantom{0}8.8} & \gray{12.1} & \gray{12.2} \\
       & \gray{doc only} & \gray{46.1} & \gray{27.5} & \gray{35.7} & \gray{89.6} & \gray{41.6} & \gray{18.3} & \gray{11.8} & \gray{15.1} & \gray{41.0} \\
       $\textsc{PEGASUS}$ & temp+doc & 63.9 & 44.9 & 50.4 & 93.0 & 67.6 & 36.8 & 28.5 & 32.6 & 40.6 \\ 
       & \gray{temp only} & \gray{54.4} & \gray{34.1} & \gray{41.4} & \gray{91.8} & \gray{\textbf{75.7}} & \gray{30.6} & \gray{\phantom{0}7.9} & \gray{19.3} & \gray{19.8} \\
       & \gray{doc only} & \gray{47.0} & \gray{28.2} & \gray{36.2} & \gray{89.8} & \gray{41.2} & \gray{18.8} & \gray{13.0} & \gray{15.9} & \gray{42.2} \\
       $\textsc{T5}$ & temp+doc & \textbf{67.0} & \textbf{48.6} & \textbf{53.4} & \textbf{93.5} & 70.9 & 43.1 & 30.5 & 36.8 & 40.6 \\ 
       & \gray{temp only} & \gray{54.4} & \gray{33.6} & \gray{40.6} & \gray{91.7} & \gray{75.1} & \gray{27.4} & \gray{\phantom{0}7.6} & \gray{17.5} & \gray{16.5} \\
       & \gray{doc only} & \gray{47.2} & \gray{29.0} & \gray{37.0} & \gray{90.0} & \gray{42.5} & \gray{18.2} & \gray{12.5} & \gray{15.3} & \gray{\textbf{42.4}} \\
    \bottomrule
    \end{tabular}
    \vspace{-2mm}
    \caption{\textbf{R}OUGE-$\{1,2,\text{L}\}$, \textbf{B}ERT\textbf{S}core, \textbf{C}EAF-\textbf{R}EE $\text{F}_1$ scores, and NLI metrics (see \S\ref{sec:evaluation}) on the \mucsum test set. ChatGPT and GPT-4 results are zero-shot and reflect averages across three prompts. BART, PEGASUS, and T5 are fine-tuned and reflect averages across three training runs. Ablation results are in \gray{gray}.\vspace{-5mm}}
    \label{tab:main_results}
    
\end{table*}

\section{Evaluation}
\label{sec:evaluation}
Apart from our human evaluation (\S\ref{sec:discussion}), we report a standard suite of automatic summarization metrics, including ROUGE-$\{1,2,\text{LCS}\}$ \citep{lin-2004-rouge, lin-och-2004-automatic} and BERTScore \cite{zhang-etal-2019-bertscore}. Since EKS summaries focus on event participants, we also report the CEAF-REE $\text{F}_1$ metric of \citet{du-etal-2021-grit}, a form of argument $\text{F}_1$ for string-fill roles. We train the neural span extractor of \citet{xia-etal-2021-lome} to extract and type arguments of the five entity-valued roles from the \mucsum summaries. We then use the trained extractor to extract arguments from each generated summary and report CEAF-REE $\text{F}_1$ scores for these extracted arguments relative to those in the gold templates.

Finally, recent metrics based on Natural Language Inference (NLI) entailment ($\rightarrow$) probabilities have been shown to exhibit higher correlation with human summary quality judgments than prior metrics \citep{chen-eger-2023-menli}. Letting $\langle S_p, S_r \rangle$ denote a predicted-reference summary pair for document $D$, we report the following probabilities: $S_p \rightarrow S_r$, $S_r \rightarrow S_p$, $S_p \leftrightarrow S_r$, and $D \rightarrow S_p$.\footnote{Entailment classification is 3-way, so probabilities $>\frac{1}{3}$ indicate predicted entailment. Further details in \autoref{app:models}.}

\section{Experiments}
\label{sec:experiments}
\subsection{Fine-Tuning}
\paragraph{Setup}
We fine-tune several standard pretrained language models (LMs) for EKS: BART \citep{lewis-etal-2020-bart}, PEGASUS \citep{zhang-etal-2020-pegasus}, and T5 \citep{raffel-etal-2020-t5}, using the large versions of each. We provide as input the document concatenated with a linearized representation of the template and its arguments. We train on the gold summaries for 30 epochs, using ROUGE-1 scores on the dev split to select the best checkpoint. For inference, we use beam search decoding with a beam size of 5 and constrain the summary length to be no longer than the length of the longest summary in the training data (256 tokens). For inputs that exceed the context window size ($W$), we right-truncate the document text. Additional details on models and inputs are provided in \autoref{app:models}.

\paragraph{Results}
Average test set metrics across three training runs for each model are given in the bottom three \texttt{temp+doc} rows of \autoref{tab:main_results}. T5 maintains a slight-to-moderate edge over the other two models on most metrics, with BART exhibiting particularly competitive (and, for \textbf{CR} and $S_p \rightarrow S_r$, superior) performance. Prior work has noted the tendency of PEGASUS to produce more extractive summaries \citep{ladhak-etal-2023-pre}, which may in part explain why it underperforms the other two on our abstractive benchmark. All three models tend to produce summaries that are entailed both by the reference ($S_r \rightarrow S_p$) and the document ($D \rightarrow S_p$).\footnote{We also calculated $D \rightarrow S_r = 0.410$ on the test set, suggesting that BART and PEGASUS are approaching human-level scores on $D \rightarrow S_p$.}

\subsection{Ablations}
\paragraph{Setup} One might wonder how much the summaries in \mucsum actually synthesize information from \emph{both} the document and query event. Most structure-to-text tasks, such as AMR-to-text \citep[][\emph{i.a.}]{pourdamghani-etal-2016-generating, flanigan-etal-2016-generation} and SQL-to-text \citep[][\emph{i.a.}]{koutrika-2010-explaining, iyer-etal-2016-summarizing} condition generation \emph{only} on the relevant structured representation (the AMR graph or the SQL query). Conversely, traditional summarization conditions only on the input document. These two setups provide natural baselines against which to compare the results discussed so far, which condition on both the document and query event. As such, we consider an ablation of the fine-tuned models in which we provide as input either \emph{only} the event template or \emph{only} the document.

\paragraph{Results} The results of these experiments are in the \texttt{temp only} and \texttt{doc only} rows of \autoref{tab:main_results}. Across most metrics, we observe degradations when ablating either the document or the template from the input, strongly indicating that \mucsum summaries \emph{do} generally leverage both the document and the event template. In most cases, this degradation is more severe when ablating the document, which makes intuitive sense, as the summaries are deliberately written to be targeted to the event represented by the \emph{template}. The superior performance of \texttt{doc only} on $D \rightarrow S_p$ and of \texttt{temp only} on $\textbf{CR}$ are intelligible when considering that templates are not needed to generate \emph{some} summary that is entailed by the document, nor is the document needed to generate \emph{some} string that contains all the template's arguments. Yet, both are necessary for a maximally informative, contextualized summary ($\textbf{R}_{1,2,L}$, \textbf{BS}, $S_r \leftrightarrow S_p$).

\subsection{Zero-Shot Prompting}
\paragraph{Setup} Finally, we present zero-shot prompted results using ChatGPT and GPT-4 \citep{openai-2023-gpt4}. To avoid inflating scores, we use three \emph{different} prompts from the one used to generate candidate summaries for \mucsum annotation, and report average results across the three, using the same prompts for both models.\footnote{A modest effort was invested in manually identifying effective prompts using several training examples, but we leave a thorough prompt engineering study for future work.}

\paragraph{Results}
Results are in the second and third rows of \autoref{tab:main_results}. While $\textbf{R}_{1,2,L}$, \textbf{BS}, and \textbf{CR} scores trail those of the fine-tuned models, the NLI metrics are actually superior in a number of cases. This may indicate that these models are capable of generating decent summaries, but ones with much less lexical overlap with the references than the summaries the from fine-tuned models exhibit.

\section{Analysis}
\label{sec:discussion}
Lastly, we consider how the summaries generated by the above models actually fare under human evaluation. We solicited summary quality judgments from three fluent English speakers on 30 randomly selected $\langle D, \langle E, R \rangle\rangle$ pairs from the test split. For each pair, annotators provided a single, five-point Likert-scale quality judgment for the summary generated for that pair by each of the five models in \autoref{tab:main_results}, plus the reference summary.\footnote{We use the summaries associated with the checkpoint/prompt that obtained the highest dev ROUGE-1 score.} Annotators were given information about the event ontology and were asked to consider the following attributes (in order of importance) in making their judgments: \emph{factuality}, \emph{adequacy}, \emph{coherence}, \emph{relevancy}, and \emph{fluency}. Annotators were not shown the source of each summary, and summary presentation order was randomized across examples.

\begin{figure}
    \centering
    \includegraphics[width=\columnwidth]{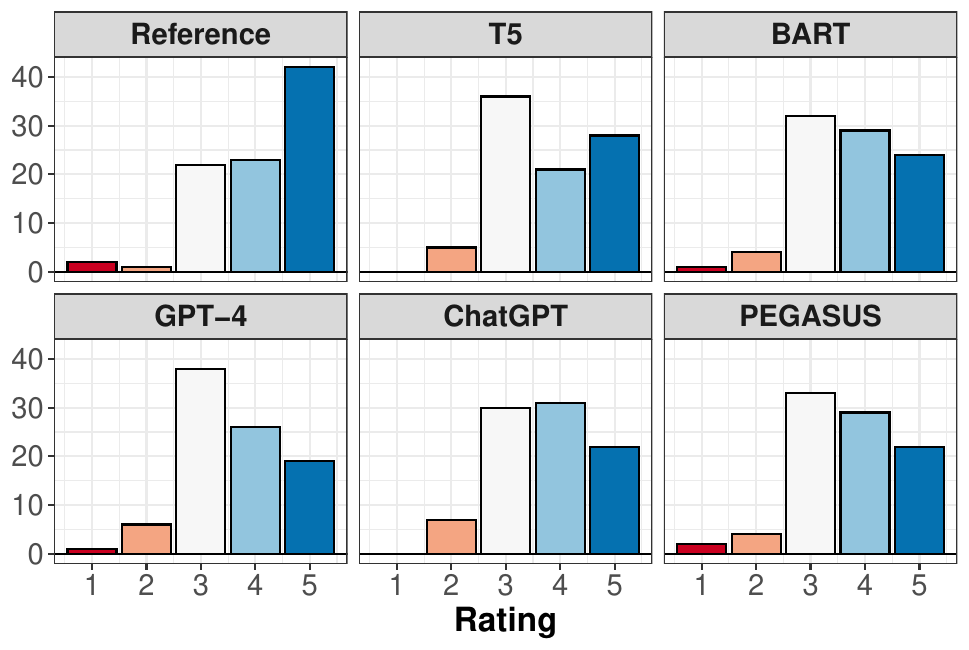}
    \vspace{-10mm}
    \caption{Distribution of ratings for models' summaries across 30 documents in quality evaluation.}
    \vspace{-5mm}

    \label{fig:quality-eval-histogram}
\end{figure}

\autoref{fig:quality-eval-histogram} shows the distribution of responses for each model's summaries. To compare these ratings, we conducted paired Wilcoxon rank sum tests for each pair of models, computing the difference between the rating an annotator gave for a particular model's summary for a particular $\langle D, \langle E, R \rangle\rangle$ against each other model's summary for that $\langle D, \langle E, R \rangle\rangle$. 

We find that all models produce summaries that are reliably worse than the reference ($p$s $< 0.01$): the best model outputs are rated 0.33 points worse than the reference on average (BART), with some models yielding an average difference of as much as 0.51 (GPT-4). The differences among models are generally much smaller: all are less than 0.2, with GPT-4 and PEGASUS tending to perform worse than BART and T5 (consistent with the results in \S\ref{sec:experiments}), though no differences are reliable ($p$s $> 0.1$). \autoref{app:quality_eval} contains further discussion.


\section{Conclusion}
\label{sec:conclusion}
We have introduced the task of \emph{event-keyed summarization} (EKS), in which the goal is to generate a summary of a \emph{specific} target event described in a document, given an underlying event ontology. We have introduced a robust benchmark for EKS, \mucsum, and presented a suite of fine-tuned and zero-shot baseline results across a diverse array of metrics. Our ablations reveal that \mucsum effectively synthesizes targeted event information with its document context. Lastly, our human evaluation testifies to the quality of the reference summaries, while showing that our baselines also yield summaries of reasonable quality.

\section*{Limitations}
\label{sec:limitations}
Although \mucsum is a valuable resource for EKS, it is narrow both in terms of its subject matter and in terms of the coverage of its ontology. While this makes it useful as an initial case study for EKS, other document-level EE datasets that have more diverse or sophisticated ontologies, such as FAMuS \citep{vashishtha2023famus} or MAVEN \citep{wang2023maven}, may require longer and more detailed summaries, and performance on such ontologies may be worse than we observe here.

Additionally, in the interest of focusing specifically on summarization quality, we generate summaries exclusively from gold templates in this work. However, in most real-world scenarios, these would be replaced by system-extracted templates, which would likely yield a degradation in summary quality proportional to the performance of the system. Experiments that consider the full extraction-to-generation pipeline would thus be an intriguing avenue for follow-up work.

\section*{Ethics}
\label{sec:ethics}
While the MUC-4 dataset has an established history in the NLP and IE communities, the documents it contains---and our \mucsum summaries, by extension---concern historical incidents of terrorism and use the names of real persons involved in them. As such, caution is obviously warranted in using this data in the training, development, or deployment of models for EKS or any other task. Given the fallibility of summarization models, it is possible and even likely that models trained on this data will make inaccurate statements concerning these historical incidents and others. We intend MUCSUM to be used for academic purposes only.

\section*{Acknowledgements}
This work was supported by NSF-BCS (2040831). Alexander Martin was also supported in part by the River Campus Libraries and the Goergen Institute for Data Science at the University of Rochester.

\bibliography{anthology,custom}
\bibliographystyle{acl_natbib}

\clearpage
\appendix

\section{Model and Implementation Details}
\label{app:models}
\subsection{OpenAI Models}
Candidate summaries were generated using the \texttt{gpt-3.5-turbo} model, accessed via the OpenAI Chat API on November 25, 2023.\footnote{\url{https://platform.openai.com/docs/guides/text-generation/chat-completions-api}} Default Chat API hyperparameters were used, with the following exceptions: (1) temperature was set to 0.8; (2) the maximum number of new tokens (i.e.\ tokens in the summary) was set to 256. The OpenAI Chat API allows users to specify both \emph{system prompts}, which provide high-level instructions about the task to be performed, as well as \emph{user prompts}, which generally provide the data to be operated on. For summary creation, we supply the following as the system prompt for all examples:

\begin{quote}
    \small{I will give you a document and a bulleted list of information about an event that the document describes. Using AT MOST 3 sentences, I want you to generate a short, accurate summary that includes ALL the information I provide you in the list. Additionally, please include information about the time and location of the attack if it is given in the document. You absolutely CANNOT include any other information that is not provided in the list. DO NOT include any extraneous details. DO NOT use more than 3 sentences.}
\end{quote}

For each example, the user prompt has the format shown below. Text between angle brackets ($\langle \texttt{text} \rangle$) is a placeholder, populated with the relevant value for each target example. Text between square brackets ([\texttt{text}]) is included only if a template has a non-null value for that slot:

\begin{quote}
    \small{Document: $\langle \texttt{document text} \rangle$
    
    Event Information:
    \begin{itemize}
        \item Event Type: $\langle \texttt{event type} \rangle$
        \item Stage of Execution: $\langle \texttt{StageOfExecution} \rangle$
        \item $[$Individual Perpetrators: $\langle \texttt{perp1}, \ldots, \texttt{perpN} \rangle]$
        \item $[$Organizations Responsible: $\langle \texttt{org1}, \ldots, \texttt{orgN} \rangle]$
        \item $[$Weapons: $\langle \texttt{weapon1}, \ldots, \texttt{weaponN} \rangle]$
        \item $[$Victims: $\langle \texttt{victim1}, \ldots, \texttt{victimN} \rangle]$
        \item $[$Physical Targets: $\langle \texttt{target1}, \ldots, \texttt{targetN} \rangle]$
    \end{itemize}
    Summary:}
\end{quote}

The three user prompts used for the zero-shot results in \S\ref{sec:experiments} are the same as above, but we vary the system prompt, using three different paraphrases of the system prompt above. For ChatGPT, we again use the \texttt{gpt-3.5-turbo} model and for GPT-4, we use the \texttt{gpt4} model, both with the same API query parameters as listed above. These experiments were run on December 8, 2023.

\subsection{Fine-Tuned Models}
Training and inference of our fine-tuned models was done with the HuggingFace Transformers (v4.35.2) and Tokenizers (v0.15.0) Python (v3.10.13) libraries \citep{wolf-etal-2019-huggingface}. We use the \texttt{t5-large} (770M params), \texttt{facebook/bart-large} (406M params), and \texttt{google/pegasus-large} (568M params) pretrained checkpoints available on the HuggingFace Hub (v0.19.4). We did not perform any hyperparameter search on these models, relying on the reasonable defaults provided by the HuggingFace API. We additionally rely on the default inference (``generation'') configuration for each model, with the exception of a uniform beam size (5) across all three models and the constraints on the minimum (15) and maximum (256) number of generated tokens. We used 1337, 1338, and 1339 as the random seeds for the training runs for each of the three models. We train each model on a single NVIDIA RTX 6000 GPU.

For the \texttt{temp+doc} setting, the input for each example consists of the document text, followed by a linearized representation of the template:
$$\langle \text{document} \rangle [\textsc{SEP}] \langle \text{template} \rangle$$
with \textsc{BOS} and \textsc{EOS} tokens inserted as required by the specific model. Drawing inspiration from \citet{du-etal-2021-grit}, the $\langle \text{template} \rangle$ representation uses a special role-delimiter token, $[\textsc{RSEP}]$, chosen from each model's additional special token vocabulary, to delineate fillers for different roles, and also includes a description of the role:
\begin{align*}
&[\textsc{RSEP}]\langle\text{role 1 description}\rangle: \langle\text{role 1 value(s)}\rangle\\
&\ldots \\
&[\textsc{RSEP}]\langle\text{role $N$ description}\rangle: \langle\text{role $N$ value(s)}\rangle
\end{align*}
Where $\langle \text{role } i \text{ value(s)} \rangle$ is a comma-separated list of role fillers and where the role descriptions are (in order), \emph{event type}, \emph{completion}, \emph{date}, \emph{location}, \emph{individual perpetrators}, \emph{organizations responsible}, \emph{physical targets}, \emph{victims}, \emph{weapons}. For the entity-valued roles, a single mention is used to represent each entity. For the \texttt{doc only} ablation, only $\langle \text{document} \rangle$ is used. For the \texttt{temp only} ablation, only $\langle \text{template} \rangle$ is used. When the input exceeds the context window length ($W = 1,024$ for all models), only the document text is truncated, and it is truncated right-to-left.

\subsection{NLI Metrics}
For our NLI metrics, we use the \texttt{menli} Python package released by \citet{chen-eger-2023-menli}.\footnote{\url{https://github.com/cyr19/MENLI}} We use the entailment probability ($e$) alone, following the authors' observation that it generally yields reasonable results (in lieu of the other formulas they consider that incorporate the neutral ($n$) and contradiction ($c$) probabilities), and we do not mix the NLI metrics with any others (i.e.\ we set \texttt{nli\_weight}=1.0). We use NLI-R as the underlying model, which is a RoBERTa-large model \citep{liu-etal-2019-roberta} fine-tuned on several NLI datasets. The metrics we report in the main text cover both reference-based ($S_p \rightarrow S_r$, $S_r \rightarrow S_p$, $S_p \leftrightarrow S_r$) and reference-free ($D \rightarrow S_p$) settings. $S_p \leftrightarrow S_r$ is simply the mean of $S_p \rightarrow S_r$ and $S_r \rightarrow S_p$.

\subsection{Other Metrics}
We use the implementations of ROUGE-\{1,2,LCS\} and BERTScore provided by the HuggingFace Evaluate (v0.4.1) Python library.\footnote{\url{https://huggingface.co/docs/evaluate}} For CEAF-REE (\textbf{CR}), we use a lightly adapted version of the implementation provided by \citet{du-etal-2021-template} that excludes the event type from the micro-average scores.\footnote{\url{https://github.com/xinyadu/gtt}. The \texttt{Location} and \texttt{Date} roles are also excluded.} We train the span extraction system of \citet{xia-etal-2021-lome} on \mucsum, using RoBERTa-large \citep{liu-etal-2019-roberta} as the encoder.\footnote{\url{https://hub.docker.com/r/hltcoe/lome}} We report exact span match (matching span boundaries and matching slot type) and partial span match (matching span boundaries, ignoring slot type) metrics in \autoref{tab:sf_results}. For both, we obtain $\text{F}_1$ scores in the low-to-mid 70s. While these are strong scores, they are not perfect and our \textbf{CR} results should be interpreted cautiously---as those for any model-based metric should be. As a final note, \textbf{CR} for the gold test set summaries is 78.2, which puts the best models on this metric in \autoref{tab:main_results} (the \texttt{temp only} ablations) within several points of human-level performance.

\begin{table}[h]
    \centering
    \begin{tabular}{l|ccc}
    \toprule
         & \textbf{P} & \textbf{R} & $\textbf{F}_1$ \\
    \midrule
        Exact Span Match & 72.5 & 75.2 & 73.8 \\
        Partial Span Match & 73.8 & 76.6 & 75.2 \\
    \bottomrule
    \end{tabular}
    \caption{Exact and partial span match P/R/$\text{F}_1$ of our span extraction system on the \mucsum test split.}
    \label{tab:sf_results}
\end{table}

\subsection{Preprocessing}
As the MUC-4 data does not have canonical sentence splits, we use the SpaCy (v3.7.2) sentence tokenizer to obtain sentence boundaries and their default word-level tokenizer for English to obtain the statistics used in \autoref{tab:dataset_statistics}.\footnote{\url{https://spacy.io/}}

\section{Annotation Agreement}
\label{app:annotation}
Instructions for the \mucsum summary annotation are included in the GitHub repository. As we note in \S\ref{sec:data}, a single annotator wrote the summary for each $\langle D, \langle E, R \rangle \rangle$ pair. However, in the interest of providing some measure of inter-annotator agreement, all annotators annotated the same random sample of 30 documents from the test split. In \autoref{tab:agreement}, we report a subset of the metrics from \autoref{tab:main_results} on these annotations---alternately treating the summaries of one annotator as the ``reference'' and those of the other two as ``predictions.''

Perhaps the most important observation is that, across metrics, there are numerous cases (i.e.\ metric-annotator pair combinations) in which one can find a superior result from one of the models in \autoref{tab:main_results}---though with the important caveat that these scores are not calculated on the same items. This offers some testament to the strength of our baselines (and of the fine-tuned models in particular). At the same time, it suggests that ``human-level'' summarization performance perhaps sits lower on the scales of these metrics than we may reflexively be inclined to think, and that numbers higher than these should not automatically be read as better. Across the NLI metrics, for instance, the highest entailment score we observe is 53.0 (between A1 and A3 on $S_p \rightarrow S_r$) which---though better than any NLI result in \autoref{tab:main_results}---is still far from 100. We thus echo the many calls from this literature to be wary of any individual summarization metric \citep[][\emph{i.a.}]{bhandari-etal-2020-evaluating, deutsch-etal-2021-statistical, gehrmann-etal-2023-repairing}, but we do not think this warrants their dismissal (see \autoref{app:quality_eval}).

\begin{table*}[]
    \small{}
    \centering
    \begin{tabular}{cc|cccccccc}
    \toprule
        $S_r$ & $S_p$ & $\textbf{R}_1$ & $\textbf{R}_2$ & $\textbf{R}_L$ & $\textbf{BS}$ & $S_r \rightarrow S_p$ & $S_p \rightarrow S_r$ & $S_r \leftrightarrow S_p$ & $D \rightarrow S_r$\\
    \midrule
         \multirow{2}{*}{A1} & A2 & 57.0 & 40.8 & 49.3 & 92.1 & \phantom{0}3.4 & 46.8 & 25.1 & \multirow{2}{*}{48.9} \\
         & A3 & 53.3 & 36.4 & 41.8 & 91.3 & 13.2 & 53.0 & 33.1 & \\
    \midrule
         \multirow{2}{*}{A2} & A1 & 57.0 & 40.8 & 49.3 & 92.1 & 46.8 & \phantom{0}3.4 & 25.1 & \multirow{2}{*}{50.2}\\
         & A3 & 71.1 & 53.1 & 59.0 & 94.0 & 36.4 & 49.7 & 43.1 \\
    \midrule
         \multirow{2}{*}{A3} & A1 & 53.3 & 36.4 & 41.8 & 91.3 & 53.1 & 13.2 & 33.1 & \multirow{2}{*}{46.1} \\
         & A2 & 71.1 & 53.0 & 59.0 & 94.0 & 49.7 & 36.4 & 43.1 & \\
    \bottomrule
    \end{tabular}
    \caption{Agreement among the authors (annotators of \mucsum) on 30 test set examples, as measured by the metrics reported in the main text. The annotator in the $S_r$ column is treated as the ``reference'' and the annotator in the $S_p$ column is treated as the ``prediction'' (but note that the distinction is moot for all metrics except $S_r \rightarrow S_p$ and $S_p \rightarrow S_r$). Also note that these are \emph{not} the same annotators as in \autoref{tab:correlations}. See discussion in \autoref{app:annotation}.}
    \label{tab:agreement}
\end{table*}

\section{Human Evaluation}
\label{app:quality_eval}
The participants in our human evaluation study (\S\ref{sec:discussion}) comprised three fluent English-speaking volunteers (unpaid), two of whom are students at the authors' home institution and one of whom is a personal acquaintance of one of the authors. The task was designed to be doable in under two hours. Annotators were permitted to ask clarificatory questions about the task, but the authors provided no input on particular judgments. Full data from the study, along with instructions, is included in the supplementary materials.
\subsection{Example}
Below, we include an example document from the evaluation study, along with the candidate summaries and their annotated quality scores (in consistent annotator order). MUC-4 documents are uncased, but we have added casing to the document and summaries below for readability.

\noindent\textbf{Document:}
\begin{quote}
    The Government of El Salvador strongly condemns the terrorist attacks carried out with explosives today against the Committee of Mothers of Political Prisoners and Missing in El Salvador and against the Salvadoran Workers National Union Federation. As a result of these attacks, several persons were wounded and others died. The government offers its condolences. The government repudiates these irrational actions and the terrorist attack carried out on 30 October by the FMLN against the Armed Forces staff installations, where 1 civilian died and 14 others were wounded. It is evident that these actions are part of a plot whose objective is to block the dialogue initiated by President Alfredo Cristiani. The government has ordered a careful investigation to determine what happened and to file charges against criminal groups that have unleashed this increased wave of terrorist violence to create a climate of unrest and to block the road to peace. Finally, the Government of El Salvador reaffirms its determination to continue the search for peace through the dialogue, and it states that in spite of these bloody events, Salvadorans must remain calm and not give up hopes to find a solution to the conflict through civilized means.

    31 October 1989, National Secretariat of Communications.
\end{quote}

\noindent\textbf{Template:}
\begin{itemize}
\setlength\itemsep{.1em}
    \item \texttt{Event Type}: Attack
    \item \texttt{StageOfCompletion}: Accomplished
    \item \texttt{Date}: 30 October
    \item \texttt{Location}: El Salvador
    \item \texttt{PerpInd}: criminal groups
    \item \texttt{PerpOrg}: FMLN
    \item \texttt{Target}: Armed Forces staff installations
    \item \texttt{Victim}: 1 civilian, 14 others
    \item \texttt{Weapon}: -
\end{itemize}
\noindent\textbf{Summaries}:
\paragraph{Reference} On 30 October, the FMLN, one of the criminal groups in El Salvador, carried out a terrorist attack on Armed Forces staff installations in El Salvador, resulting in the death of 1 civilian and injuries to 14 others.
\noindent\textbf{Scores}: 5, 5, 3

\paragraph{BART} The Government of El Salvador strongly condemns the terrorist attack carried out on 30 October by the FMLN against the Armed Forces staff installations, where one civilian died and 14 others were wounded. It is evident that these actions are part of a plot to block the dialogue initiated by President Alfredo Cristiani.
\noindent\textbf{Scores}: 4, 3, 4

\paragraph{PEGASUS} On 30 October, the FMLN carried out a terrorist attack on the Armed Forces staff installations in El Salvador, where one civilian died and 14 others were wounded.
\noindent\textbf{Scores}: 4, 3, 3

\paragraph{T5} The Government of El Salvador strongly condemns the terrorist attacks carried out with explosives on 30 october by criminal groups affiliated with the FMLN against the Armed Forces staff installations, where one civilian died and 14 others were wounded.
\noindent\textbf{Scores}: 2, 4, 3

\paragraph{ChatGPT} On October 30, 1989, in El Salvador, a terrorist attack was carried out by the FMLN against the Armed Forces staff installations, resulting in the death of one civilian and the injury of 14 others. The Government strongly condemned these actions, which were part of a plot to block the dialogue initiated by President Alfredo Cristiani. A careful investigation has been ordered to determine what happened and file charges against the criminal groups responsible for this increased wave of terrorist violence.
\noindent\textbf{Scores}: 5, 5, 4

\paragraph{GPT-4} On October 30, 1989, the FMLN, along with associated criminal groups, carried out a terrorist attack on Armed Forces staff installations in El Salvador. the attack resulted in the death and injury of several civilians. This event, which was part of a broader scheme to disrupt President Alfredo Cristiani's initiated dialogue, has been condemned by the Salvadoran Government who are conducting an investigation.
\noindent\textbf{Scores}: 2, 5, 2

\subsection{Judgment-Metric Correlations} Here, we consider the extent to which the judgments from the human evaluation of the model and reference summaries (\S\ref{sec:discussion}) correlate with automatic metrics. In \autoref{tab:correlations}, we report Kendall's rank correlation coefficient (two-sided $\tau$-c) between each evaluator's quality judgments on the 180 summaries (30 examples $\times$ 6 candidate summaries/example) and the corresponding metric value for that item, for each metric in \autoref{tab:main_results}.

Intriguingly, the only reliably positive correlations ($p$s $< 0.05$) we observe are for annotators B1 ($\textbf{R}_{1,2,L}, \textbf{BS}, \textbf{CR}$) and B2 ($\textbf{R}_{1,2,L}, \textbf{BS}, D \rightarrow S_p$). Of particular interest is that, \emph{contra} the findings of \citet{chen-eger-2023-menli}, we observe almost no reliable positive correlations among the NLI metrics (B2 on $D \rightarrow S_p$ excepted). One part of the explanation very likely lies in the difference between our dataset and the ones they study, which include SummEval \citep{fabbri-etal-2021-summeval}, RealSum \citep{bhandari-etal-2020-evaluating}, and Rank19 \citep{falke-etal-2019-ranking}---all of which have corpora focused on very different topics from  MUC-4/\mucsum, and which relied on somewhat different (and differently prioritized) evaluation criteria for their judgments.

Another part of the explanation may lie in the fact that \citeauthor{chen-eger-2023-menli} used different entailment-based ``formulas'' in their summarization results, depending on which performed best on a particular dataset and setting (reference-based vs.\ reference-free evaluation). We use \emph{one} of these formulas in this work (the entailment probability, $e$), whereas they further consider others that incorporate the neutral ($n$) and contradiction ($c$) probabilities, such as $-c$, $e-n$, and $e-c$.

It is also worth noting that BERTScore ($\textbf{BS}$) is at least not obviously superior in our study to ROUGE ($\textbf{R}$), \emph{contra} findings from \citet{zhang-etal-2019-bertscore}---though again, differences in the data and judgment task may help account for this.

Even so, we do not think this is cause for a wholesale dismissal of automatic metrics. Each of the metrics considered here \emph{does} convey information about a candidate summary that is likely to be useful in real-world contexts (e.g.\ degree of lexical overlap is very informative for plagiarism detection)---it is simply \emph{different} information from what is captured by a human judgment. Moreover, the fact that individual \emph{human} judgments can clearly exhibit such variability also suggests that, while they may (must) remain the gold standard for summarization, any \emph{particular} judgment ought to be understood as being nothing more than that.

\begin{table*}[]
    \small{}
    \centering
    \adjustbox{max width=\linewidth}{
    \begin{tabular}{c|ccccccccc}
    \toprule
        Annotator & $\textbf{R}_1$ & $\textbf{R}_2$ & $\textbf{R}_L$ & $\textbf{BS}$ & \textbf{CR} & $S_r \rightarrow S_p$ & $S_p \rightarrow S_r$ & $S_r \leftrightarrow S_p$ & $D \rightarrow S_p$\\
    \midrule
         B1 & $\phantom{-}.136^*_{\tiny{(.04)}}$ & $\phantom{-}.156^*_{\tiny{(.02)}}$ & $\phantom{-}.169^*_{\tiny{(.01)}}$ & $.127^*_{\tiny{(.05)}}$ & $\phantom{-}.057_{\tiny{(.37)}}$ & $\phantom{-}.003_{\tiny{(.97)}}$ & $-.030_{\tiny{(.65)}}$ & $\phantom{-}.004_{\tiny{(.95)}}$ & $.085_{\tiny{(.19)}}$ \\
         B2 & $\phantom{-}.208^*_{\tiny{(.00)}}$ & $\phantom{-}.211^*_{\tiny{(.00)}}$ & $\phantom{-}.182^*_{\tiny{(.01)}}$ & $.232^*_{\tiny{(.00)}}$ & $\phantom{-}.218^*_{\tiny{(.00)}}$ & $\phantom{-}.009_{\tiny{(.90)}}$ & $\phantom{-}.123_{\tiny{(.07)}}$ & $\phantom{-}.056_{\tiny{(.41)}}$ & $.152^*_{\tiny{(.02)}}$ \\
         B3 & $-.014_{\tiny{(.81)}}$ & $-.053_{\tiny{(.34)}}$ & $-.064_{\tiny{(.25)}}$ & $.001_{\tiny{(.98)}}$ & $-.030_{\tiny{(.58)}}$ & $-.023_{\tiny{(.68)}}$ & $-.033_{\tiny{(.55)}}$ & $-.021_{\tiny{(.70)}}$ & $.084_{\tiny{(.13)}}$ \\
    \midrule
         Avg & $\phantom{-}.114_{\tiny{(.21)}}$ & $\phantom{-}.118_{\tiny{(.09)}}$ & $\phantom{-}.113_{\tiny{(.08)}}$ & $.102_{\tiny{(.26)}}$ & $\phantom{-}.130_{\tiny{(.25)}}$ & $\phantom{-}.088_{\tiny{(.84)}}$ & $-.006_{\tiny{(.48)}}$ & $\phantom{-}.021_{\tiny{(.72)}}$ & $.119_{\tiny{(.09)}}$ \\
    \bottomrule
    \end{tabular}
    }
    \caption{Kendall's rank correlation coefficient ($\tau_{(\text{p-val})}$) between each human evaluator's judgments and the corresponding automatic metric across the 180 judgments (30 examples $\times$ 6 candidate summaries) from the human evaluation study in \S\ref{sec:discussion}. ``Avg'' indicates the macro-average correlation across evaluators. ``$^*$'' denotes significance at $p=0.05$. Note that these are \emph{not} the same annotators as in \autoref{tab:agreement}. See discussion in \autoref{app:quality_eval}.}
    \label{tab:correlations}
\end{table*}

\end{document}